\documentclass[10pt,twocolumn,letterpaper]{article}

\pdfoutput=1

\usepackage[pagenumbers]{cvpr} % To force page numbers, e.g. for an arXiv version

\usepackage{graphicx}
\usepackage{amsmath}
\usepackage{amssymb}
\usepackage{booktabs}

\usepackage{fdsymbol}
\usepackage{mathrsfs}
\usepackage{amsfonts,amssymb}
\usepackage{multirow}

\usepackage[pagebackref,breaklinks,colorlinks]{hyperref}

\usepackage[capitalize]{cleveref}
\crefname{section}{Sec.}{Secs.}
\Crefname{section}{Section}{Sections}
\Crefname{table}{Table}{Tables}
\crefname{table}{Tab.}{Tabs.}

\def\cvprPaperID{8833} % *** Enter the CVPR Paper ID here
\def\confName{CVPR}
\def\confYear{2022}

\begin{document}

\title{\textbf{Understanding and Testing Generalization of Deep Networks\\on Out-of-Distribution Data}}

\author{Rui Hu, Jitao Sang, Jinqiang Wang, Rui Hu, Chaoquan Jiang\\
    School of Computer and Information Technology, Beijing Jiaotong University \\
    {\tt\small \{ruihoo,jtsang,jinqiangwang,ritahu,cqjiang\}@bjtu.edu.cn}
}
\maketitle

\begin{abstract}
     Deep network models perform excellently on In-Distribution (ID) data, but can significantly fail on Out-Of-Distribution (OOD) data. While developing methods focus on improving OOD generalization, few attention has been paid to evaluating the capability of models to handle OOD data. This study is devoted to analyzing the problem of experimental ID test and designing OOD test paradigm to accurately evaluate the practical performance. Our analysis is based on an introduced categorization of three types of distribution shift to generate OOD data. Main observations include: (1) ID test fails in neither reflecting the actual performance of a single model nor comparing between different models under OOD data. (2) The ID test failure can be ascribed to the learned marginal and conditional spurious correlations resulted from the corresponding distribution shifts. Based on this, we propose novel OOD test paradigms to evaluate the generalization capacity of models to unseen data, and discuss how to use OOD test results to find bugs of models to guide model debugging.
\end{abstract}

%%%%%%%%% BODY TEXT
\section{Introduction}
\label{sec:intro}

\begin{figure}[!tb]
  \centering
  \begin{subfigure}{0.48\linewidth}
    \centering
    \includegraphics[width=0.75\linewidth]{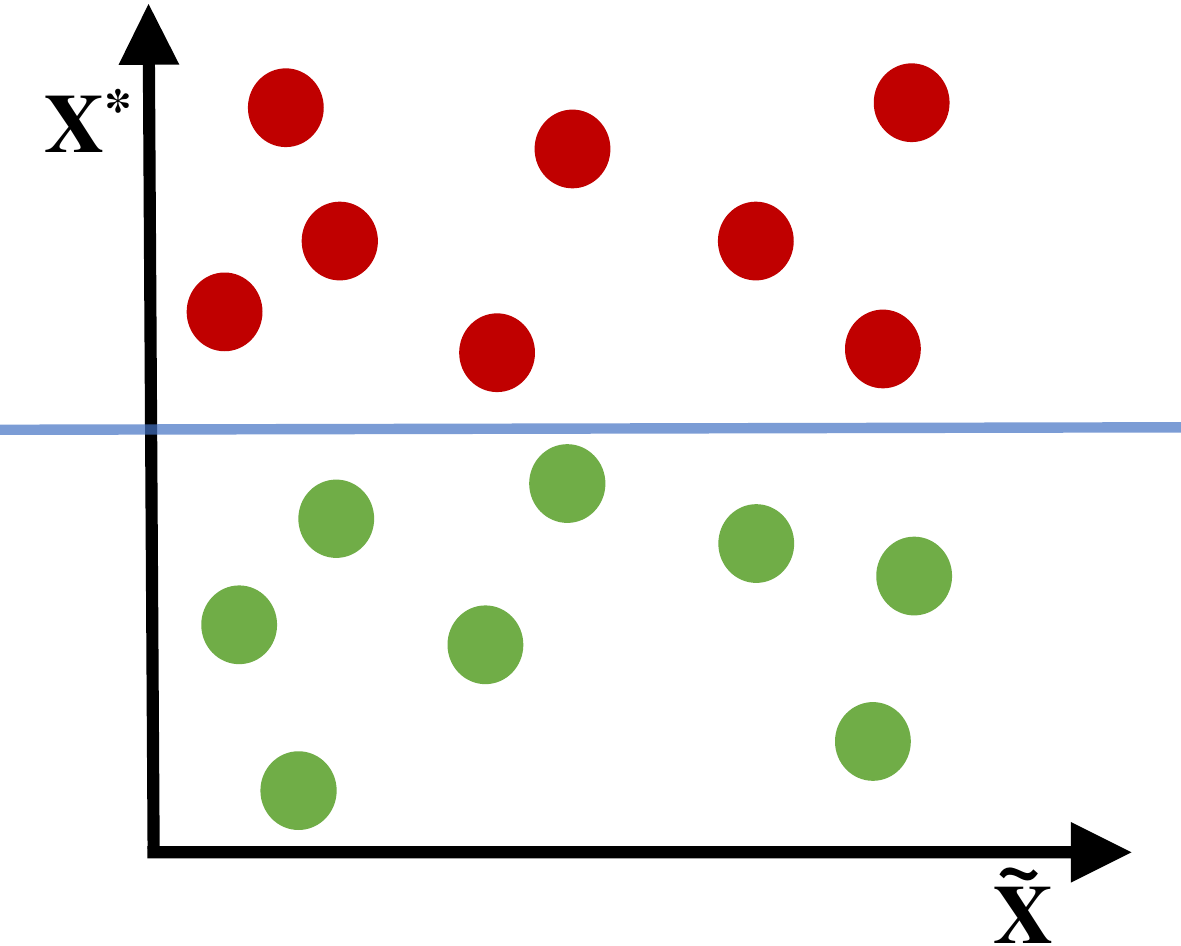}
    \caption{real distribution}
    \label{fig:ds-a}
  \end{subfigure}
  \begin{subfigure}{0.48\linewidth}
    \centering
    \includegraphics[width=0.75\linewidth]{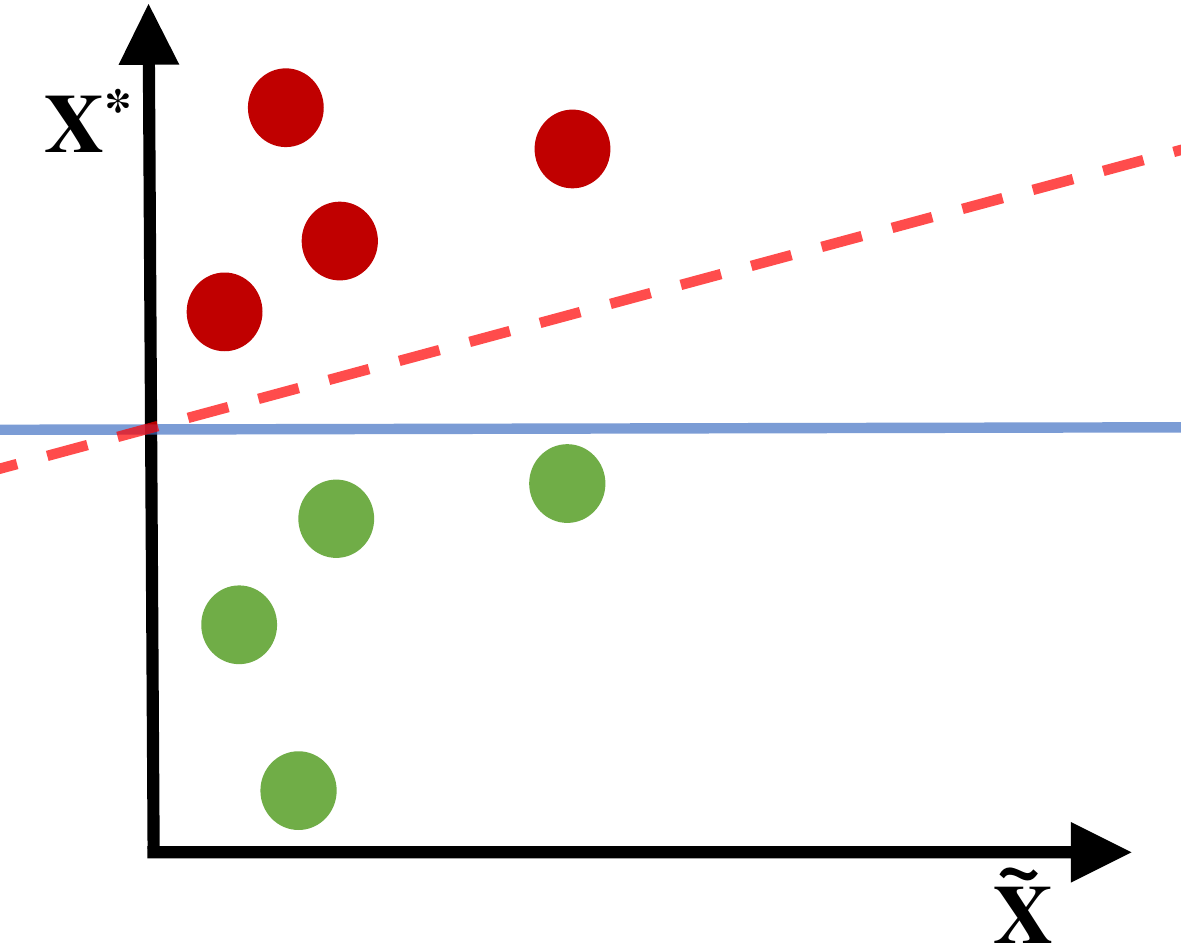}
    \caption{marginal distribution shift}
    \label{fig:ds-b}
  \end{subfigure}

  \begin{subfigure}{0.48\linewidth}
    \centering
    \includegraphics[width=0.75\linewidth]{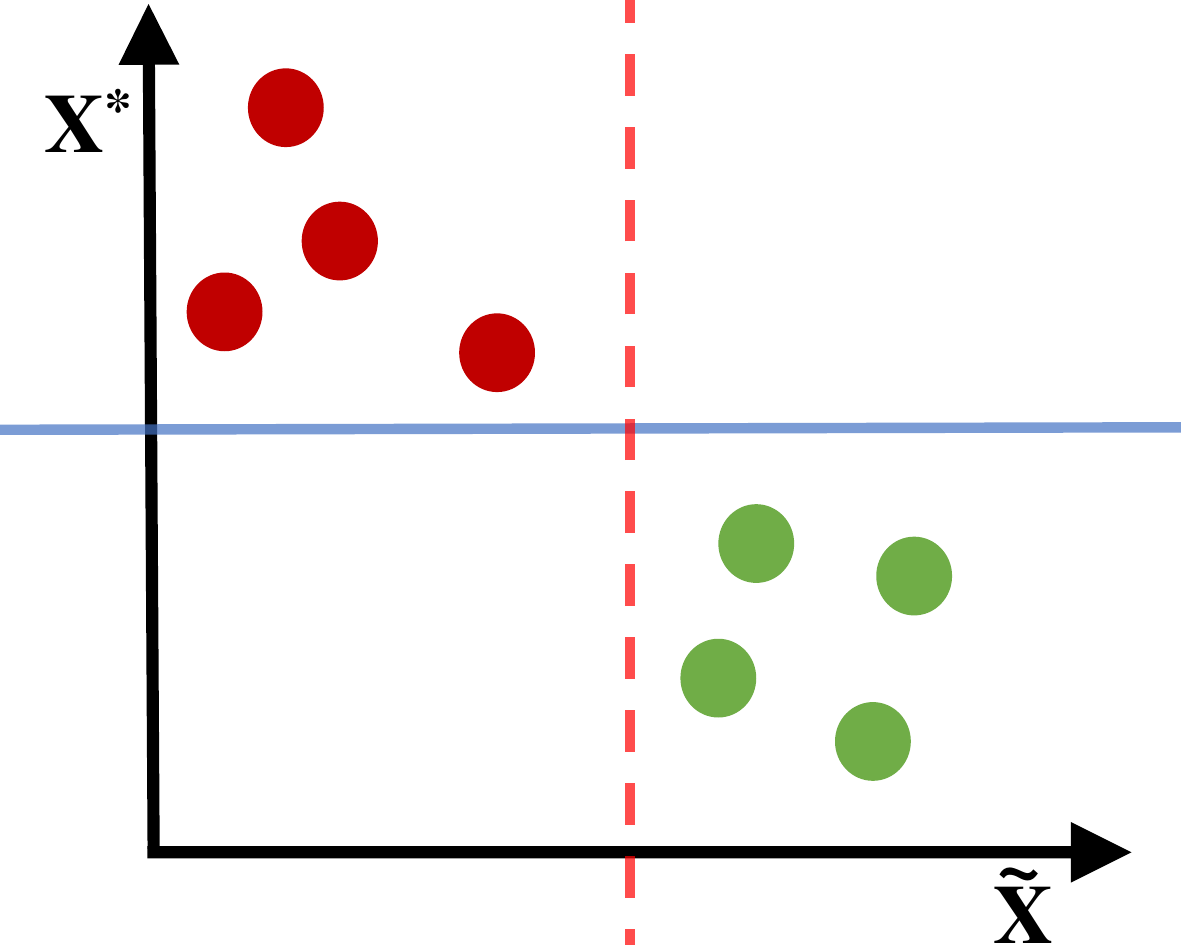}
    \caption{conditional distribution shift}
    \label{fig:ds-c}
  \end{subfigure}
  \begin{subfigure}{0.48\linewidth}
    \centering
    \includegraphics[width=0.75\linewidth]{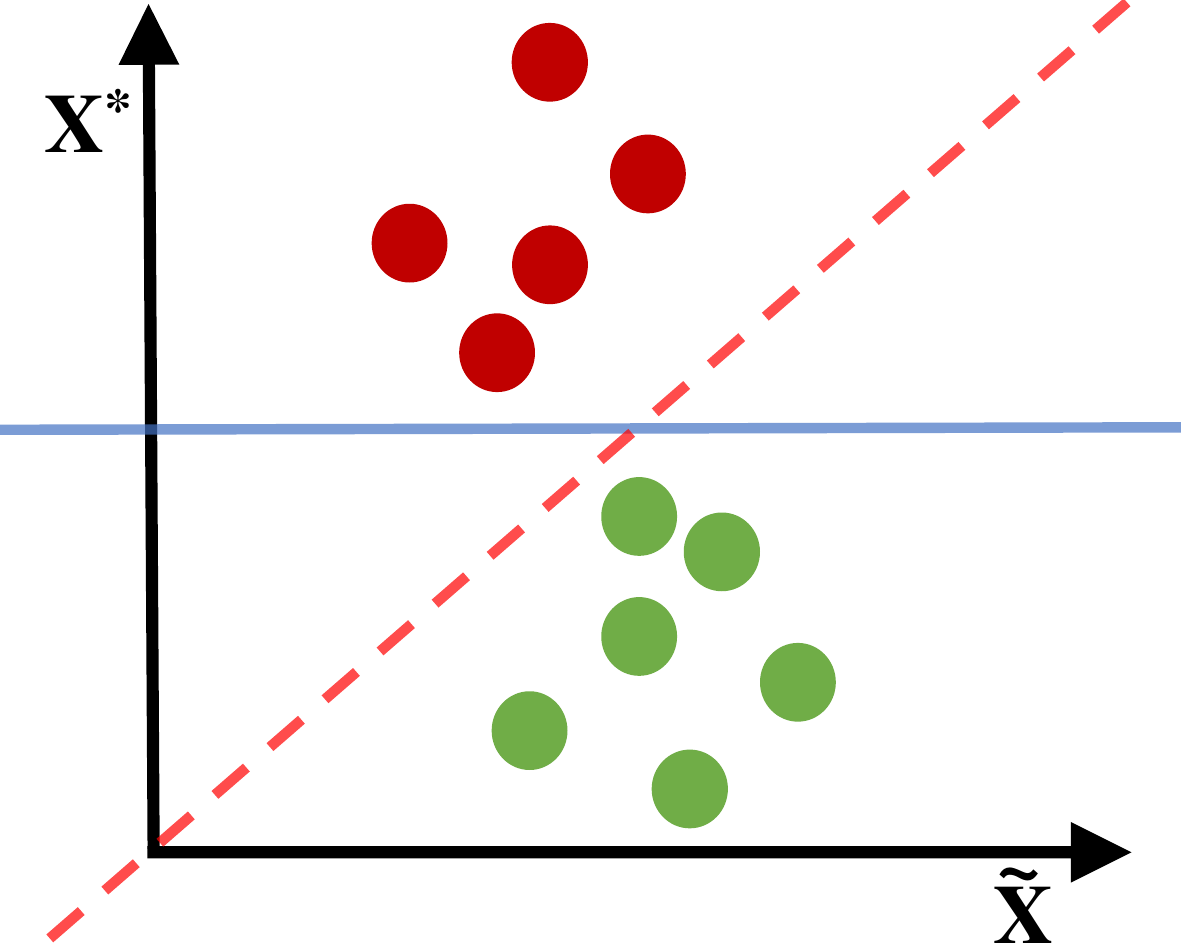}
    \caption{joint distribution shift}
    \label{fig:ds-d}
  \end{subfigure}

   \caption{Illustration of three types of \textbf{distribution shift} to generate OOD data. $X^*$ and $\widetilde{X}$ represent relevant and irrelevant features respectively. The colors of red and green dots represent two different classes. The blue solid line means the optimal decision rule and red dotted lines represent decision rules that model learn from the shifted data.}
   \label{fig:distribution shift}
\end{figure}

The last few years have witnessed human-level performance of deep network models under experimental test in various tasks like image recognition, speech processing and reading comprehension \cite{he2016deep, RajpurkarJL18squad}. However, in practical industrial implementations when the real-world data deviates from the observed training data, the model performance significantly declines and even sometimes completely fails \cite{recht2019imagenet,TaoriDSCRS20robustness,hendrycks2019benchmarking,hendrycks2021natural}. Extensive efforts have been taken to bridge the distribution gap between real-world data and training data to guarantee practical performance. Typical solutions include domain generalization \cite{ganin2015unsupervised,ganin2016domain,li2018domain,motiian2017unified}, causal learning \cite{peters2016causal,pfister2019invariant,pfister2019invariant}, stable learning \cite{shen2020stable,zhang2021deep}, etc.

Other than developing methods to improve the experiment-practical generalization capability, few attention has been paid to the issue of model testing and address questions like ``why the experimental testing results fail to reflect the real performance in practical implementation?'' We argue that the practical implementation relies on models' capability to handle Out-Of-Distribution (OOD) data, while experimental test essentially evaluates In-Distribution (ID) capability of models where training and test data follow the identical distribution. This study attempts to address the following three specific questions:
\begin{itemize}
\setlength{\parskip}{-2pt}
    \item What are the problems of ID test?
    \item Why ID test fails to reflect the practical implementation performance?
    \item How to evaluate the practical implementation performance under experimental test?
\end{itemize}

To facilitate these analyses, we first introduce a unified perspective to simulate the distribution shift between the observed training data and unseen test data. Specifically, three types of distribution shift are considered: (1) marginal distribution shift which refers to changes in the marginal distributions of irrelevant features between the training and test data; (2) conditional distribution shift which refers to changes in conditional distributions of labels when irrelevant features are given; and (3) joint distribution shift which suffers both the above situations. As illustrated in the toy example in Figure \ref{fig:distribution shift}, these distribution shifts represent typical ways to construct OOD data and mislead the model to learn non-optimal parameters for OOD generalization. Following this categorization, we construct three ID settings and the corresponding OOD settings of training and test set on CelebA \cite{liu2015deep} and Colored MNIST \cite{arjovsky2019invariant}. The analysis into the result difference between ID settings and OOD settings address the above first question and observes the following problems of ID test: (1) ID test result cannot reflect models' absolute OOD performance, (2) ID test result cannot compare between different models' relative OOD performance.

Following the observations on the problems of ID test, we further ascribe the reason that ID test fails to reflect the practical implementation performance to the spurious correlations that model learned when training. Corresponding to the marginal and conditional distribution shifts, two types of spurious correlations are discussed: (1) marginal spurious correlation that irrelevant features are employed, and (2) conditional spurious correlation that incorrect feature-label correlations are employed. Experiment validates the performance decline on OOD data resulted from the different spurious correlations. Finally, we propose practical OOD test paradigms to evaluate generalization capability of models to unseen data using the observed data. We discuss test paradigms for both scenarios when irrelevant features are known and unknown using different OOD data settings. Moreover, the derived OOD test results gain access to identify what irrelevant features critically influence the model performance (i.e., finding bugs), and provide guideline to develop targeted solutions for performance improvement (i.e., debugging). It is desired to consider the test module beyond simply evaluation and form a closed loop between testing and debugging to promote the model performance in practical implementation.

The main contributions of this paper are as follows:
\begin{itemize}
\setlength{\parskip}{-2pt}
    \item We introduce a unified perspective to simulate OOD data construction into three types of distribution shift. Experimental analysis observes the problems of ID test under corresponding distribution shift settings.
    \item We explore the reasons of ID test failure as two types of learned spurious correlations. The discussions on spurious correlations illustrate their differences in effects on the generalization capability of models.
    \item We propose novel OOD test paradigms to evaluate the models' performance in handling unseen data. This opens possibility to employ the proposed OOD test to help identify critical irrelevant features.
\end{itemize}

%%%%%%%%%%%%%%%%%%%%%%%%%%%%%%%%%%%%%%%%%%%%%%%%%%%%%%%%%%%%%%%%%%%%%%%%%%
\section{A Unified Perspective of Out-of-Distribution Data}
\label{sec:definition}

In this section, we first introduce the concepts of IID-based supervised learning and out-of-distribution data. We then propose a novel and unifying view on categorization of distribution shift. Further, we construct three ID settings and the corresponding OOD settings of training and test sets on two simple image classification datasets for experiments in Section \ref{sec:problem}.

~\\
\noindent \textbf{IID-based Supervised Learning.} Let $\mathcal{X}$ denotes the feature space and $\mathcal{Y}$ the label space.  A predictive function or model is defined as $f_{\theta}: \mathcal{X} \rightarrow \mathcal{Y}$, which takes features as input and predicted labels as output. A loss function is defined as $\ell: \mathcal{Y} \times \mathcal{Y} \rightarrow [0, +\infty)$. A data distribution is defined as $P(X,Y)$ which means a joint distribution on $\mathcal{X} \times \mathcal{Y}$. The context of supervised learning usually assumes that there exists a set of training samples $S=\{(x_i, y_i)\}_i^N$ with $(x_i, y_i)$ following $P_{tr}(X,Y)$. The goal is to find a set of optimal parameters $\Theta$ which minimize the risk $\mathbb{E}_{(x,y) \sim P_{te}}[\ell (f_{\Theta}(x), y)]$, where $P_{te}$ means test data distribution. Since we only have access to the training distribution $P_{tr}(X,Y)$ via the dataset $S$, the test distribution $P_{te}$ remains unseen to us. We instead search a predictor minimizing the empirical risk
\begin{equation}
    \mathbb{E}=\frac{1}{N} \sum_{i}^{N} \ell (f_{\Theta}(x_i), y_i),
\end{equation}
which assumes that the training samples and test samples are both independent and identically distributed (IID) \cite{vapnik1998statistical}.

~\\
\noindent \textbf{Out-of-Distribution Data.} Although the IID assumption allows us to train a model more conveniently, this assumption is difficult to be satisfied in reality. The distribution of test data can be distinct from the distribution of training data (i.e., out-of-distribution)\cite{quinonero2009dataset}, namely
\begin{equation}
    P_{tr}(X,Y) \neq P_{te}(X,Y) \label{con:2}.
\end{equation}

%%%%%%%%%%%%%%%%%%%%%%%%%%%%%%%%%%%%%%%%%%%%%%%%%%%%%%%%%%%%%%%%%%%%%%%%%%
\subsection{Three Types of Distribution Shift}

\begin{figure*}[!tb]
    \centering
    \begin{subfigure}{0.48\linewidth}
      \centering
      \includegraphics[width=0.48\linewidth,height=0.45\linewidth]{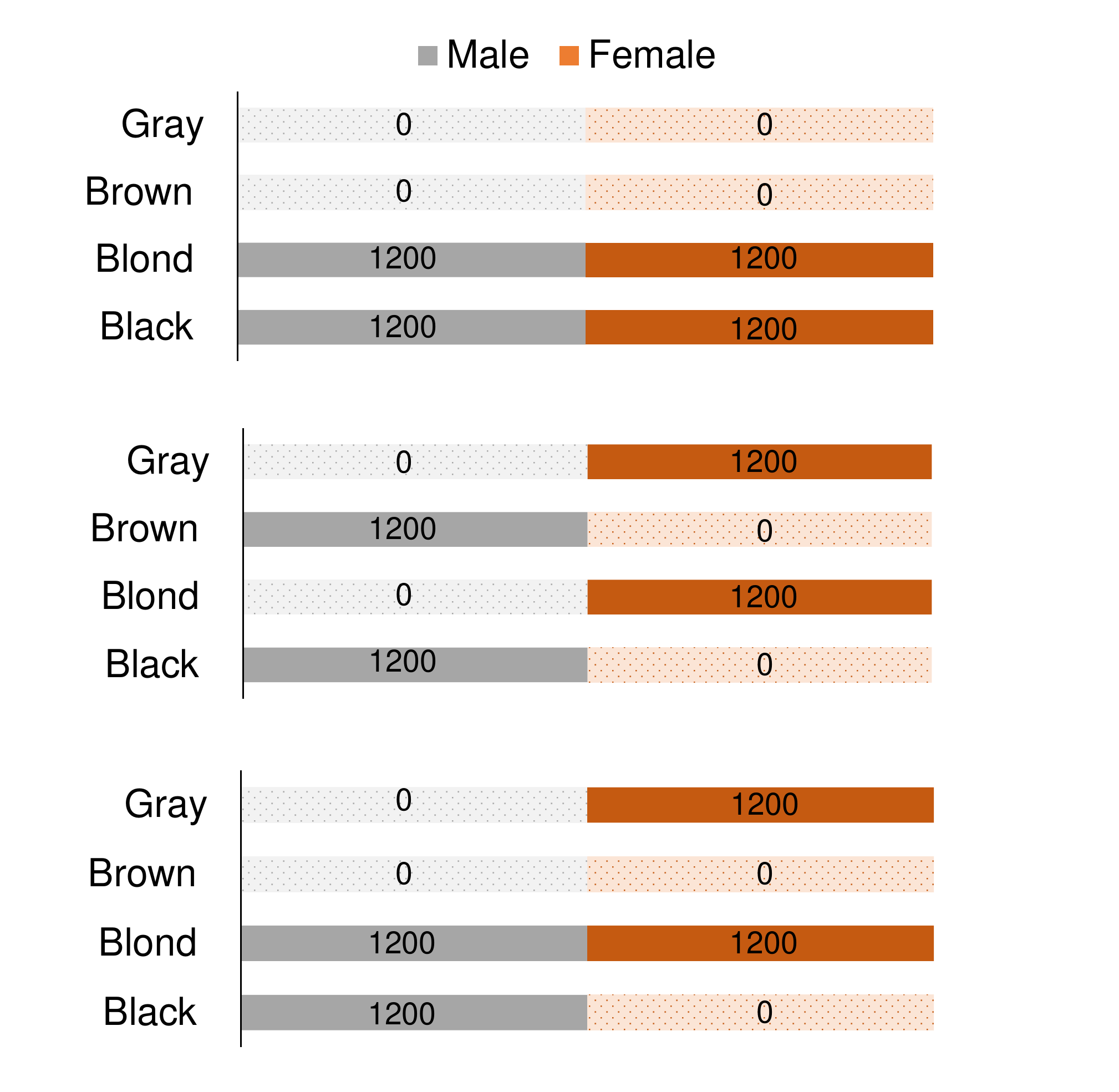}
      \includegraphics[width=0.48\linewidth,height=0.45\linewidth]{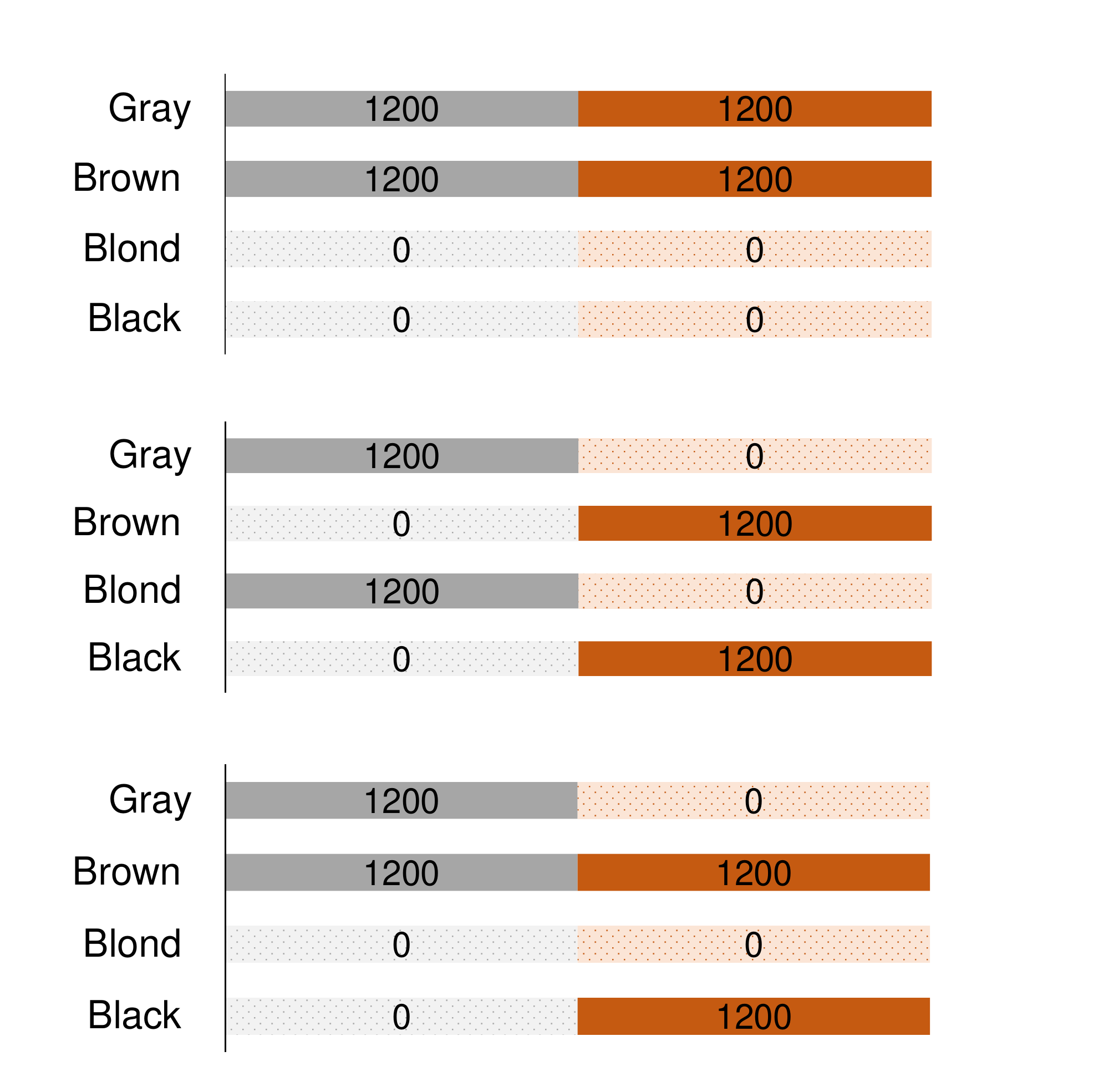}
      \caption{CelebA}
      \label{fig:celeba-set}
    \end{subfigure}
    \begin{subfigure}{0.48\linewidth}
      \centering
      \includegraphics[width=0.48\linewidth,height=0.45\linewidth]{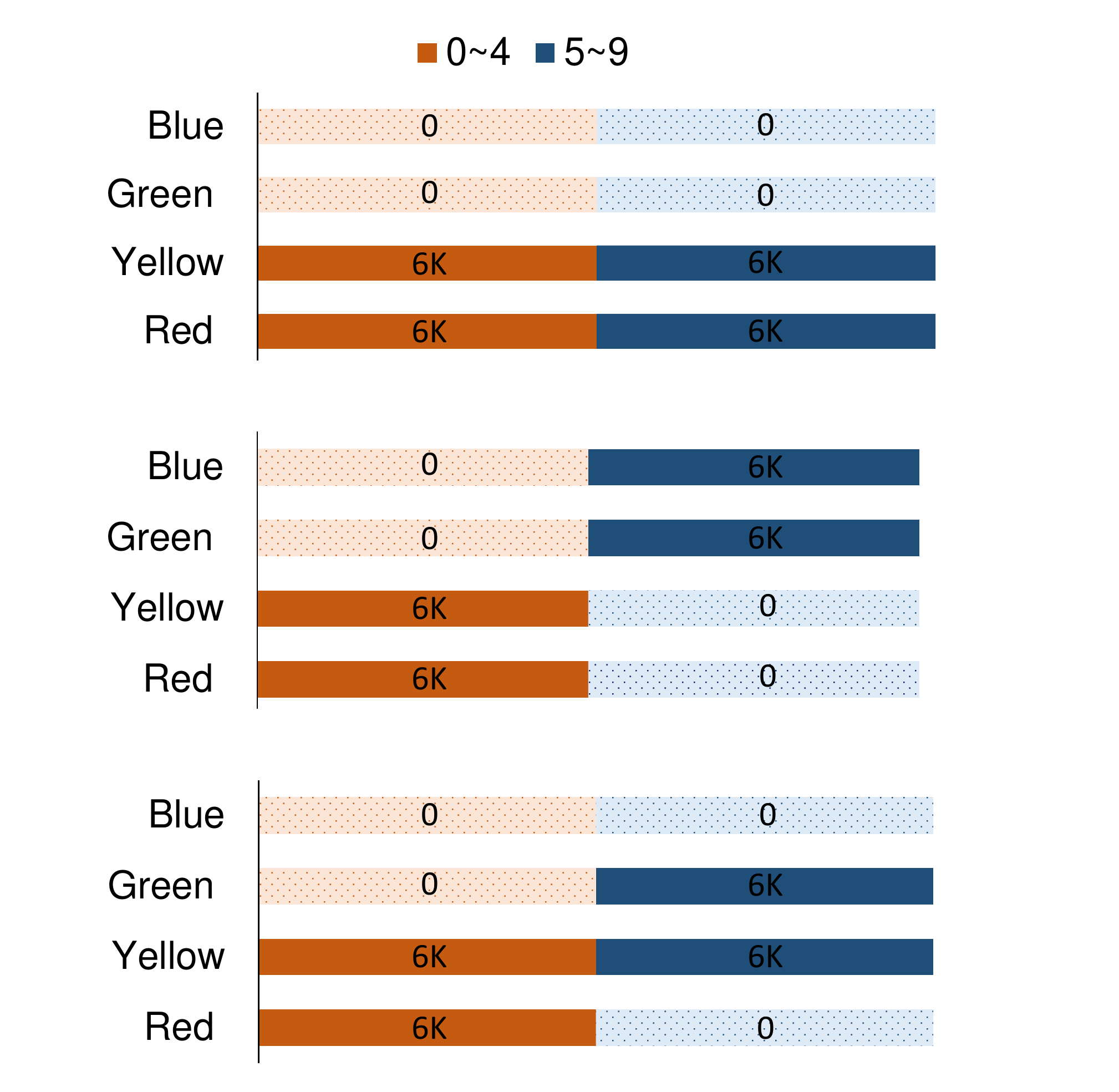}
      \includegraphics[width=0.48\linewidth,height=0.45\linewidth]{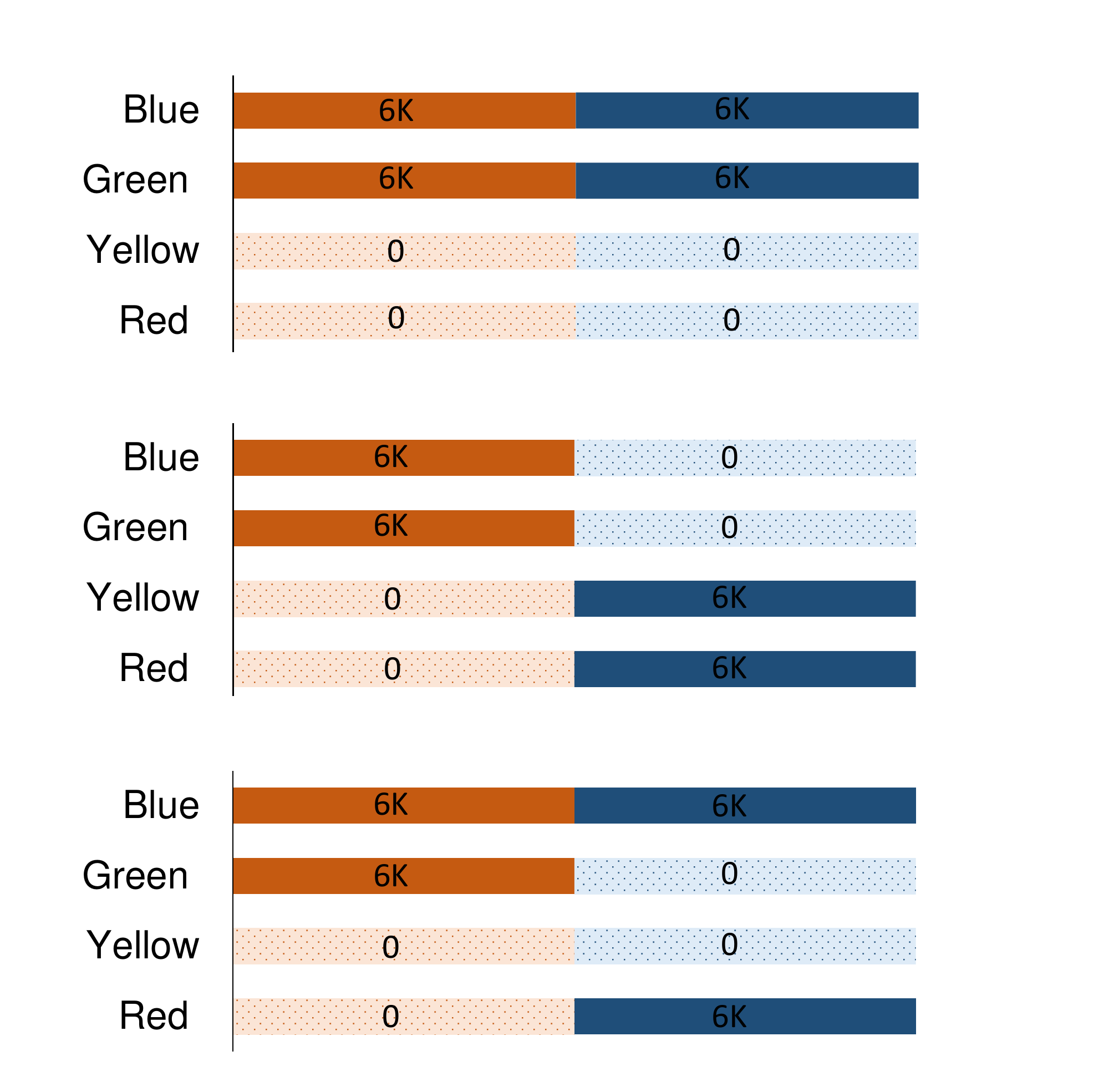}
      \caption{Colored MNIST}
      \label{fig:mnist-set}
    \end{subfigure}
    \caption{Demonstration of three types of distribution shift on CelebA and Colored MNIST. From top to bottom, there are three pairs of training and test data with different settings which represent marginal, conditional, and joint distribution shifts.}
    \label{fig:dataset-setting}
\end{figure*}

Some works assume that the cause of the distribution shift is covariate shift \cite{ben2010theory}, that is $P_{tr}(Y|X)=P_{te}(Y|X)$ and $P_{tr}(X) \neq P_{te}(X)$, of which we think as not comprehensive. In general, for most deep learning tasks, we categorize features into relevant features $X^*$ and irrelevant features $\widetilde{X}$. As shown in the Figure \ref{fig:ds-a}, the relevant features determine the real correspondence of input data and label. For example, determining which species a cat belongs to is relevant to its pattern and coat, not to its background, nor whether it is jumping or sleeping. We further make some reasonable assumptions to simplify Equation \ref{con:2}.

\noindent \textbf{Assumption 1.} Distribution of relevant features is invariant both in training and test data (e.g., all Ragdoll cats have blue eyes), which means $P(X^*, Y)$ remains unchanged \footnote{There are some works devoted to studying the situation where $P(X^*,Y)$ changes, such as concept drift \cite{gama2014conceptsurvey}, which is not included in the discussion of this work.}.

\noindent \textbf{Assumption 2.} Relevant features and irrelevant features are conditionally independent when label is given, denoted as $X^* \Vbar \widetilde{X} \mid Y$:
\begin{equation}
    \begin{aligned}
        & P(X^*,\widetilde{X}|Y)=P(X^*|Y)P(\widetilde{X}|Y).
    \end{aligned}
\end{equation}

\noindent \textbf{Assumption 3.} There is no label shift \cite{garg2020labelshift} between training data and test data, denoted as $P_{tr}(Y)=P_{te}(Y)$.

~\\
\noindent With these assumptions, we can rewrite $P(X,Y)$ as
\begin{equation}
    \begin{aligned}
        P(X^*,\widetilde{X},Y) &= \frac{P(X^*,Y)P(\widetilde{X},Y)}{P(Y)},
    \end{aligned}
\end{equation}
\noindent and the essential distribution shift between training and test data is the shift of $P(\widetilde{X},Y)$ between them. According to the joint probability formula, we have:
\begin{equation}
    P(\widetilde{X},Y) = P(\widetilde{X})P(Y|\widetilde{X}).
\end{equation}
\noindent Therefore, there are three ways to construct OOD test set, that is, distribution shift between training and test data can be divided into three types:
\begin{itemize}
\setlength{\parskip}{0pt}
    \item \textbf{1) Marginal distribution shift} on $\widetilde{X}$. As shown in Figure \ref{fig:ds-b}, $P(\widetilde{X})$ changes while $P(Y|\widetilde{X})$ remains unchanged;
    \item \textbf{2) Conditional distribution shift} on $Y$ when $\widetilde{X}$ is given. As shown in Figure \ref{fig:ds-c}, $P(Y|\widetilde{X})$ changes while $P(\widetilde{X})$ remains unchanged;
    \item \textbf{3) Joint distribution shift}, a combination of 1) and 2). $P(Y|\widetilde{X})$ and $P(\widetilde{X})$ both change, as shown in Figure \ref{fig:ds-d}.
\end{itemize}
Taking a binary classification task of dogs and cats as an example, we consider backgrounds as irrelevant features. Suppose the backgrounds have four types: \textit{in cage}, \textit{at home}, \textit{on grass} and \textit{on snow}.
If available data only cover one or two types of background, there occurs marginal distribution shift. If we get all four types of background, but the backgrounds are correlated with the labels (e.g., cats all in cages or at home, dogs all on grass or snow), a conditional distribution shift occurs. The joint distribution shift occurs when it suffers both the above situations. For example, the backgrounds of available data are not full-scale and partly correspond to labels. As depicted in Figure \ref{fig:distribution shift}, in these cases, the model can learn red lines that deviate from optimum blue ones, which achieve high accuracy on in-distribution data but cannot generalize to out-of-distribution data.

%%%%%%%%%%%%%%%%%%%%%%%%%%%%%%%%%%%%%%%%%%%%%%%%%%%%%%%%%%%%%%%%%%%%%%%%%%
\subsection{OOD Dataset Setting}

\begin{table*}[!htb]
\caption{In- and out-of-distribution accuracy of models and the decline ratio among them.}
\centering
\begin{tabular}{l l lll lll lll}
\toprule
  &  &  \multicolumn{3}{c}{Marginal distribution shift} &
        \multicolumn{3}{c}{Conditional distribution shift} &
        \multicolumn{3}{c}{Joint distribution shift} \\
\cline{3-5} \cline{6-8} \cline{9-11}
\multicolumn{1}{l}{}  &                             & ID   & OOD   & Drop\%              & ID   & OOD   & Drop\%              & ID   & OOD   & Drop\%  \\
\hline \hline
\multirow{7}{*}{\rotatebox[origin=c]{90}{CelebA}}
    & ResNet18 \cite{he2016deep}                    & 96.68 & 84.64 & $\downarrow$12.45   & 99.36 & 25.27 & $\downarrow$74.57   & 96.29 & 59.31 & $\downarrow$38.40   \\
    & AlexNet \cite{krizhevsky2012alexnet}          & 98.33 & 88.91 & $\downarrow$9.58    & 98.75 & 26.93 & $\downarrow$72.73   & 97.08 & 69.79 & $\downarrow$28.11   \\
    & Vgg11 \cite{SimonyanZ14vgg}                   & 99.02 & 91.32 & $\downarrow$7.78    & 98.70 & 33.70 & $\downarrow$65.86   & 97.40 & 71.36 & $\downarrow$26.74   \\
    & DensNet121 \cite{huang2017densenet}           & 97.27 & 89.31 & $\downarrow$8.18    & 99.80 & 24.15 & $\downarrow$75.80   & 97.66 & 71.38 & $\downarrow$26.91   \\
    & SqueezeNet1.0 \cite{iandola2016squeezenet}       & 97.96 & 86.65 & $\downarrow$11.55   & 98.74 & 22.48 & $\downarrow$77.23   & 96.07 & 63.73 & $\downarrow$33.66   \\
    & ResNeXt50 \cite{xie2017resnext}               & 97.85 & 86.04 & $\downarrow$12.07   & 99.41 & 37.52 & $\downarrow$62.26   & 97.27 & 70.31 & $\downarrow$27.72   \\
    & \textbf{Average}                                       & \textbf{97.85} & \textbf{87.81} & $\downarrow$\textbf{10.26}   & \textbf{99.13} & \textbf{28.34} & $\downarrow$\textbf{71.41}   & \textbf{96.96} & \textbf{67.65} & $\downarrow$\textbf{30.23}   \\
\hline \hline
\multirow{6}{*}{\rotatebox[origin=c]{90}{Colored Mnist}}
    & ResNet18 \cite{he2016deep}                    & 96.80 & 49.51 & $\downarrow$48.85   & 98.17 & 0.21  & $\downarrow$99.79   & 97.20 & 7.00  & $\downarrow$92.80   \\
    & SqueezeNet1.0 \cite{iandola2016squeezenet}    & 97.87 & 37.07 & $\downarrow$62.12   & 98.33 & 0.03  & $\downarrow$99.97   & 97.90 & 32.28 & $\downarrow$67.03   \\
    & SqueezeNet1.1 \cite{iandola2016squeezenet}    & 86.96 & 26.83 & $\downarrow$69.15   & 97.77 & 0.00  & $\downarrow$100.00  & 96.90 & 13.46 & $\downarrow$86.11   \\
    & ResNet34 \cite{he2016deep}                    & 98.00 & 41.21 & $\downarrow$57.95   & 98.77 & 0.40  & $\downarrow$99.60   & 98.13 & 23.20 & $\downarrow$76.36   \\
    & MobileNet v3 \cite{howard2017mobilenets}      & 89.76 & 34.85 & $\downarrow$61.17   & 95.17 & 0.78  & $\downarrow$99.18   & 93.13 & 28.39 & $\downarrow$69.52   \\
    & \textbf{Average}                                       & \textbf{93.88} & \textbf{37.90} & $\downarrow$\textbf{59.63}   & \textbf{97.64} & \textbf{0.28}  & $\downarrow$\textbf{99.71}   & \textbf{96.65} & \textbf{20.87} & $\downarrow$\textbf{78.41}   \\
\bottomrule
\end{tabular}
\label{table:ood result}
\end{table*}

\begin{figure*}[!tb]
    \centering
    \begin{subfigure}{0.32\linewidth}
      \centering
      \includegraphics[width=0.9\linewidth]{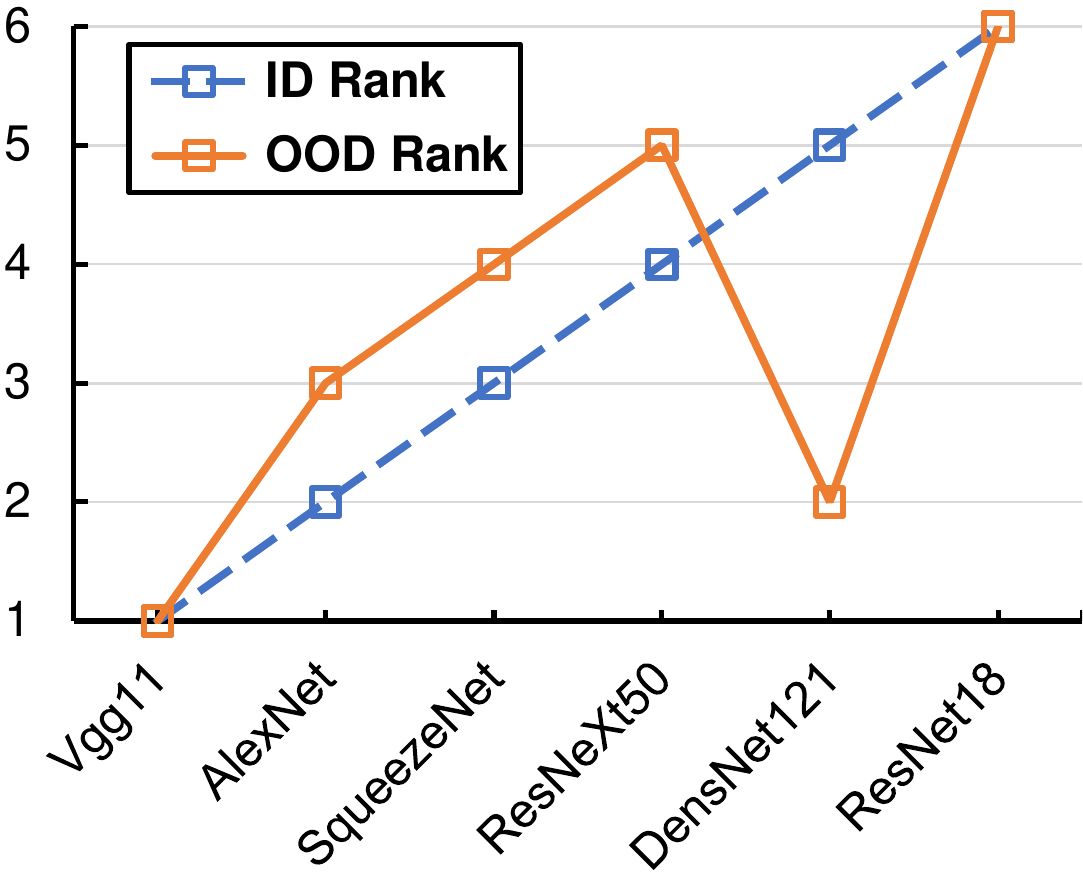}
      \caption{Marginal distribution shift}
      \label{fig:rank-a}
    \end{subfigure}
    \begin{subfigure}{0.32\linewidth}
      \centering
      \includegraphics[width=0.9\linewidth]{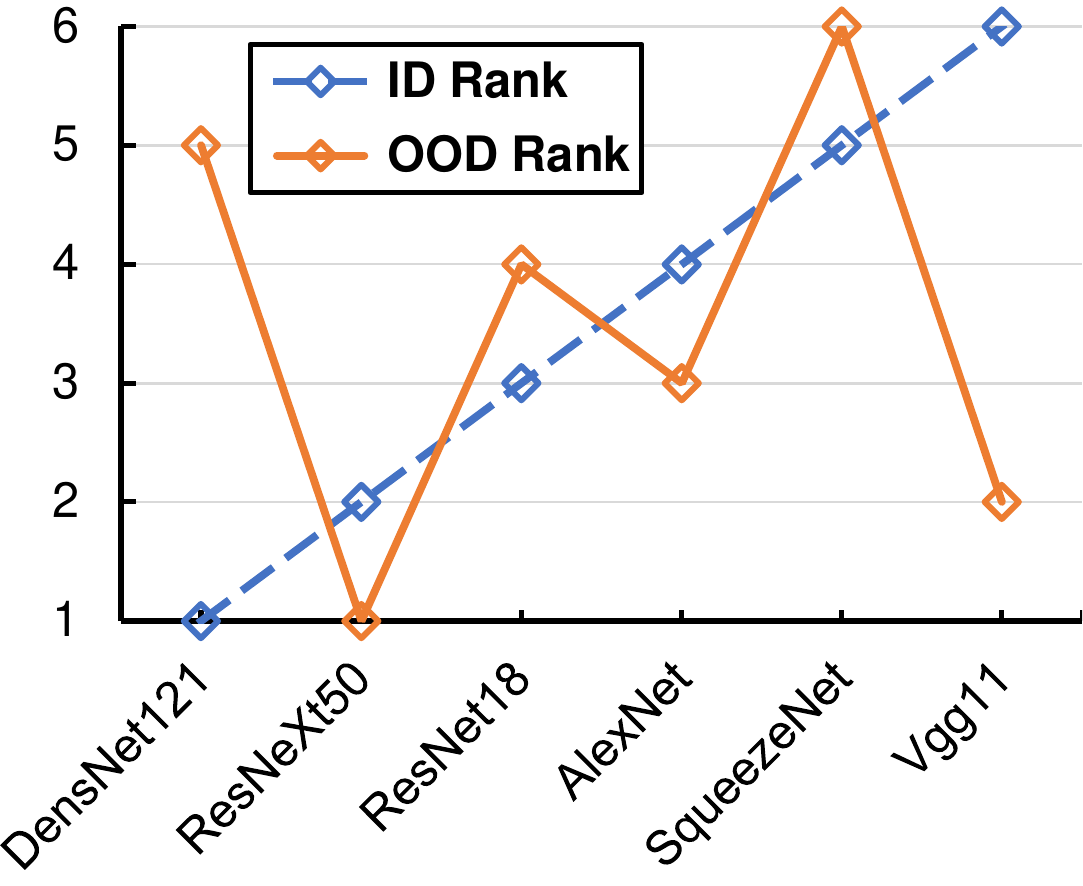}
      \caption{Conditional distribution shift}
      \label{fig:rank-b}
    \end{subfigure}
    \begin{subfigure}{0.32\linewidth}
      \centering
      \includegraphics[width=0.9\linewidth]{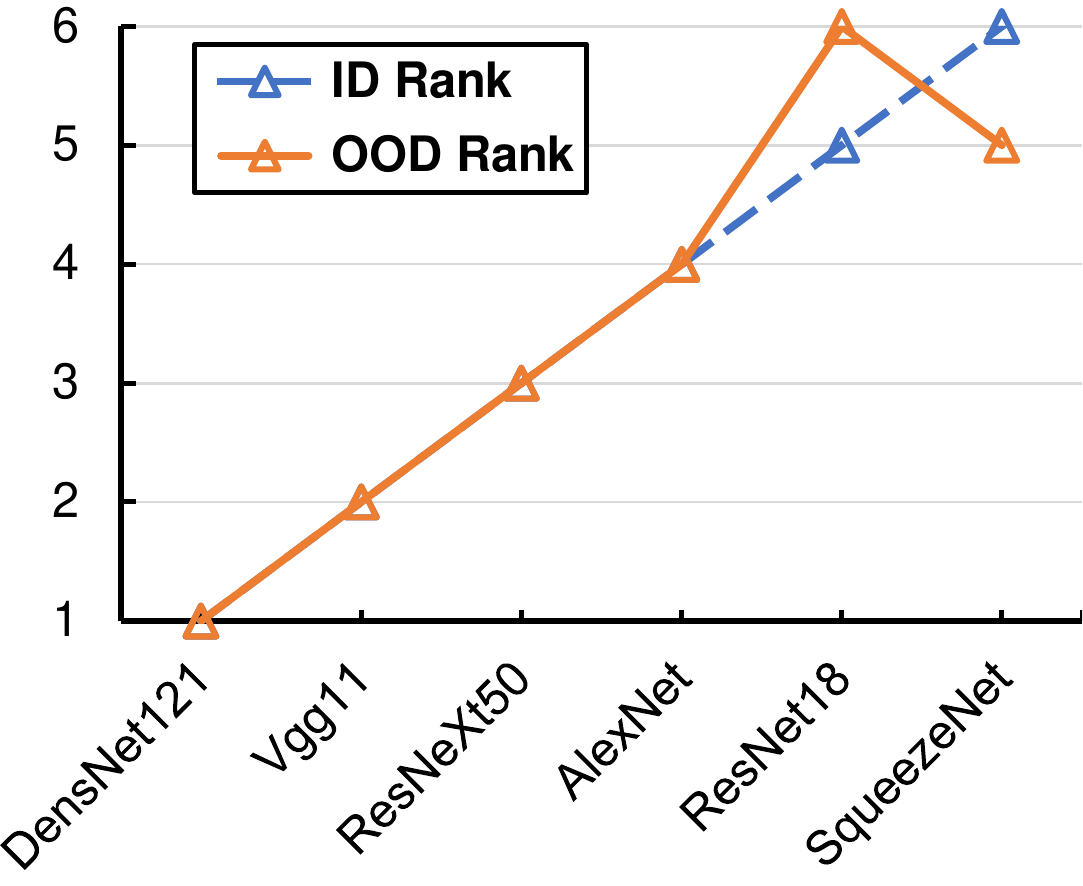}
      \caption{Joint distribution shift}
      \label{fig:rank-c}
    \end{subfigure}
    \caption{Illustration of the ID rank and OOD rank of several models under three distribution shifts on CelebA.}
    \label{fig:rank}
\end{figure*}

To investigate how models based on the IID assumption perform on OOD data, we choose two simple image classification datasets, CelebA \cite{liu2015deep} and Colored MNIST \cite{arjovsky2019invariant}, on which we construct the above three types of distribution shift.
\begin{itemize}
    \item \textbf{CelebA} \cite{liu2015deep} is a large-scale face dataset, each image with 40 attributes, and always used in Fairness area \cite{quadrianto2019discovering,wang2020towards}. In our experimental setting, we pick genders as labels and select 9,600 images in which the proportion of male and female is balanced. For either gender, there are four types of hair colors: black, blond, brown and gray, each being 1,200. Then we consider hair colors as irrelevant features and construct three pairs of dataset, each pair with a training set and an out-of-distribution test set. For instance, we select samples with blond and black hairs as training set, and make left samples test set to build marginal distribution shift. See detailed settings in Figure \ref{fig:celeba-set}.
    \item \textbf{Colored MNIST} \cite{arjovsky2019invariant} is a colored MNIST handwritten digits dataset. In our experiment, we expand the digits colors of \cite{arjovsky2019invariant} to four types: red, yellow, green, and blue. The number of each digits category is the same as the original MNIST \cite{lecun1998gradient}, and each color accounts for a quarter in each digit category, roughly 7,500 samples per color. Similar to the setting of CelebA, we construct three pairs of datasets to correspond to three distribution shifts. See detailed settings in Figure \ref{fig:mnist-set}.
\end{itemize}

%%%%%%%%%%%%%%%%%%%%%%%%%%%%%%%%%%%%%%%%%%%%%%%%%%%%%%%%%%%%%%%%%%%%%%%%%%
\section{The Problem of ID Test}
\label{sec:problem}

\begin{figure*}
    \centering
    \includegraphics[width=0.8\linewidth]{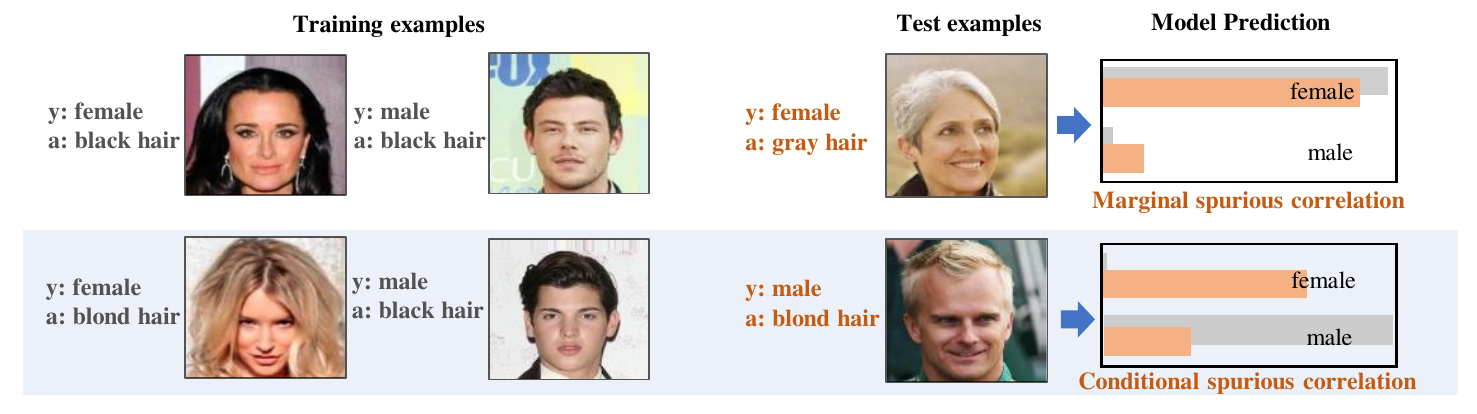}
    \caption{Examples of the two spurious correlations: (top) marginal spurious correlation; and (bottom) conditional spurious correlation. In the legend, gray and orange bars indicate probability of models’ predictions for in- and out-of-distribution samples respectively.}
    \label{fig:spurious}
\end{figure*}

One of the main goals of training deep network models is to make the models generalize in the wild. As obtaining in-the-wild test data is expensive and time-consuming, most works use ID test data, that is,  randomly splitting a portion of the training data as test data, to evaluate the models. In this section, we evaluate the performance of several ID test-based models under different distribution shifts and find several shortcomings of the ID test. See results in Table \ref{table:ood result} and Figure \ref{fig:rank}.

~\\
\noindent \textbf{Accuracy of ID Test is Not Accurate.} Similar to results of \cite{recht2019imagenet,hendrycks2021many,hendrycks-etal-2020-pretrained}, models with high accuracy on in-distribution data can fail on new distribution. From the results in Table \ref{table:ood result}, there shows a significant drop of accuracy in models when testing on OOD data. The accuracy drops range from 7.78\% to 77.23\% on CelebA and 48.85\% to 100\% on Colored MNIST. This demonstrates that the performance of the ID test cannot represent the real performance of the model, while the OOD test can reflect it to some extent. Furthermore, the models decline differently on the OOD test sets with different distribution shifts. Taking CelebA as a detailed example, the average drops of models accuracy under marginal distribution shift is 10.26\%, while under conditional and joint distribution shifts, the number is 71.41\% and 30.23\% respectively, both higher than the first. This illustrates that different distribution shifts impair models' performance in different degrees, so we need different types of OOD data for testing comprehensively. In short, the accuracy of ID test is not accurate, and we need OOD test to address this problem.

~\\
\noindent \textbf{Rank of ID Test is Not Robust.} Although ID accuracy cannot represent the real accuracy of the models, we hope that the rank of ID accuracy of models can help us pick the best one. However, in our experiment, we find that the ID rank of the models is not robust, i.e., cannot represent the OOD rank. For instance, Figure \ref{fig:rank-b} shows that under conditional distribution shift the Vgg11 model has the lowest ID rank, but a relatively high OOD rank (only lower than ResNeXt50). Some prior works show that a model with high accuracy on in-distribution data will also achieve relatively high accuracy on out-of-distribution data \cite{recht2019imagenet,miller2020effect} which is inconsistent with our finding. We think they come to this result because the distribution shift between the OOD test data and training data they used is very small. Under our dataset partition settings, the marginal and conditional distribution shifts we construct are extreme, while the joint distribution shift is less extreme. Figure \ref{fig:rank} shows that the ID ranks and OOD ranks of the models barely coincide under the marginal and conditional distribution shifts, while under the joint distribution shift the ranks almost coincide. In summary, ID test results cannot compare between different models' relative OOD ranks, and the ranks are not consistent across different types of distribution shift, so we cannot only use a single OOD test set to evaluate the real performance of models.

%%%%%%%%%%%%%%%%%%%%%%%%%%%%%%%%%%%%%%%%%%%%%%%%%%%%%%%%%%%%%%%%%%%%%%%%%%
\section{Reasons that ID Test Fails}
\label{sec:reasons}

To better guide us to generate practical OOD test set, we need to know what causes the failure of ID test. We conclude that ID test fails as a result of two different spurious correlations, which are 1) marginal spurious correlation and 2) conditional spurious correlation. These two spurious correlations incur performance drop of models under marginal and conditional distribution shifts, respectively.

\noindent \textbf{Definition 1.} Given a dataset $\mathcal{D}$ with an irrelevant feature $\widetilde{x} \in \widetilde{X}$ which makes the conditional probability of any two labels $y_i$, $y_j$ equal, namely $p(y_i|\widetilde{x})=p(y_j|\widetilde{x})$. If a model trained on $\mathcal{D}$ use $\widetilde{x}$ as a discriminative feature when making decisions, we call the model has learned a marginal spurious correlation.

\noindent \textbf{Definition 2.} Given a dataset $\mathcal{D}$ with an irrelevant feature $\widetilde{x} \in \widetilde{X}$, there exists two $y_i$, $y_j$ whose conditional probability are not equal, namely $p(y_i|\widetilde{x}) \neq p(y_j|\widetilde{x})$. If a model trained on $\mathcal{D}$ use $\widetilde{x}$ as a discriminative feature when making decisions, we call the model have learned a conditional spurious correlation.

Figure \ref{fig:spurious} shows an example, where $y$ represents the label of the image and $a$ represents the specific attribute of the irrelevant feature \textit{hair color}. The top half of the Figure \ref{fig:spurious} shows that the model trained on the dataset with all black hairs has a drop in predictions probability when testing in dataset with all gray hairs (i.e., marginal spurious correlation); at the bottom half, the gender and hair color in test data has the opposite correspondence with the training set, and the model's prediction result is also the opposite (i.e., conditional spurious correlation).

\subsection{Marginal Spurious Correlation}

\begin{table}[!t]
\centering
\caption{Results of ResNet18 and 2-layer CNN models on Colored MNIST in different contexts. \textit{R-1K} represents the dataset which size is 1,000 with red color, similar settings to other datasets in the table. The bold part represents in-distribution accuracy.}
\resizebox{\linewidth}{!}{
\begin{tabular}{ll llll c}
\toprule
 & dataset & test-R & test-Y & test-G & test-B  & Drop\%        \\
 \hline \hline
 \multirow{8}{*}{\rotatebox[origin=c]{90}{ResNet18}}
    & R-1K & \textbf{98.90}   & 27.10   & 11.60   & 9.10    & $\downarrow$62.92  \\
    & Y-1K & 34.80   & \textbf{98.80}   & 42.70   & 28.70   & $\downarrow$48.13  \\
    & G-1K & 14.40   & 19.90   & \textbf{99.10}   & 14.00   & $\downarrow$62.82  \\
    & B-1K & 23.10   & 12.40   & 18.40   & \textbf{99.10}   & $\downarrow$61.40  \\
    & R-Full & \textbf{99.95}   & 59.70   & 72.81   & 85.75   & $\downarrow$20.40  \\
    & Y-Full & 45.72   & \textbf{99.91}   & 76.34   & 30.81   & $\downarrow$36.75  \\
    & G-Full & 57.75   & 83.19   & \textbf{99.92}   & 68.43   & $\downarrow$22.61  \\
    & B-Full & 22.19   & 34.52   & 87.54   & \textbf{99.89}   & $\downarrow$38.90  \\
    \midrule
\multirow{8}{*}{\rotatebox[origin=c]{90}{2-layer CNN}}
    & R-1K & \textbf{91.90}   & 89.60   & 47.10   & 35.20   & $\downarrow$28.24  \\
    & Y-1K & 75.10   & \textbf{91.60}   & 88.60   & 27.30   & $\downarrow$22.87  \\
    & G-1K & 80.20   & 90.70   & \textbf{92.80}   & 62.00   & $\downarrow$12.26  \\
    & B-1K & 90.40   & 86.50   & 67.30   & \textbf{97.60}   & $\downarrow$12.26  \\
    & R-Full & \textbf{99.90}   & 93.16   & 42.86   & 82.95   & $\downarrow$20.20  \\
    & Y-Full & 98.62   & \textbf{99.91}   & 98.15   & 53.33   & $\downarrow$12.42  \\
    & G-Full & 14.19   & 88.74   & \textbf{99.90}   & 29.90   & $\downarrow$41.76  \\
    & B-Full & 99.23   & 89.52   & 29.00   & \textbf{99.87}   & $\downarrow$20.49  \\
\bottomrule
\end{tabular}
}
\label{table:marginal-sc}
\end{table}

\begin{figure*}[!t]
    \centering
    \begin{subfigure}{0.48\linewidth}
        \centering
        \includegraphics[width=0.8\linewidth]{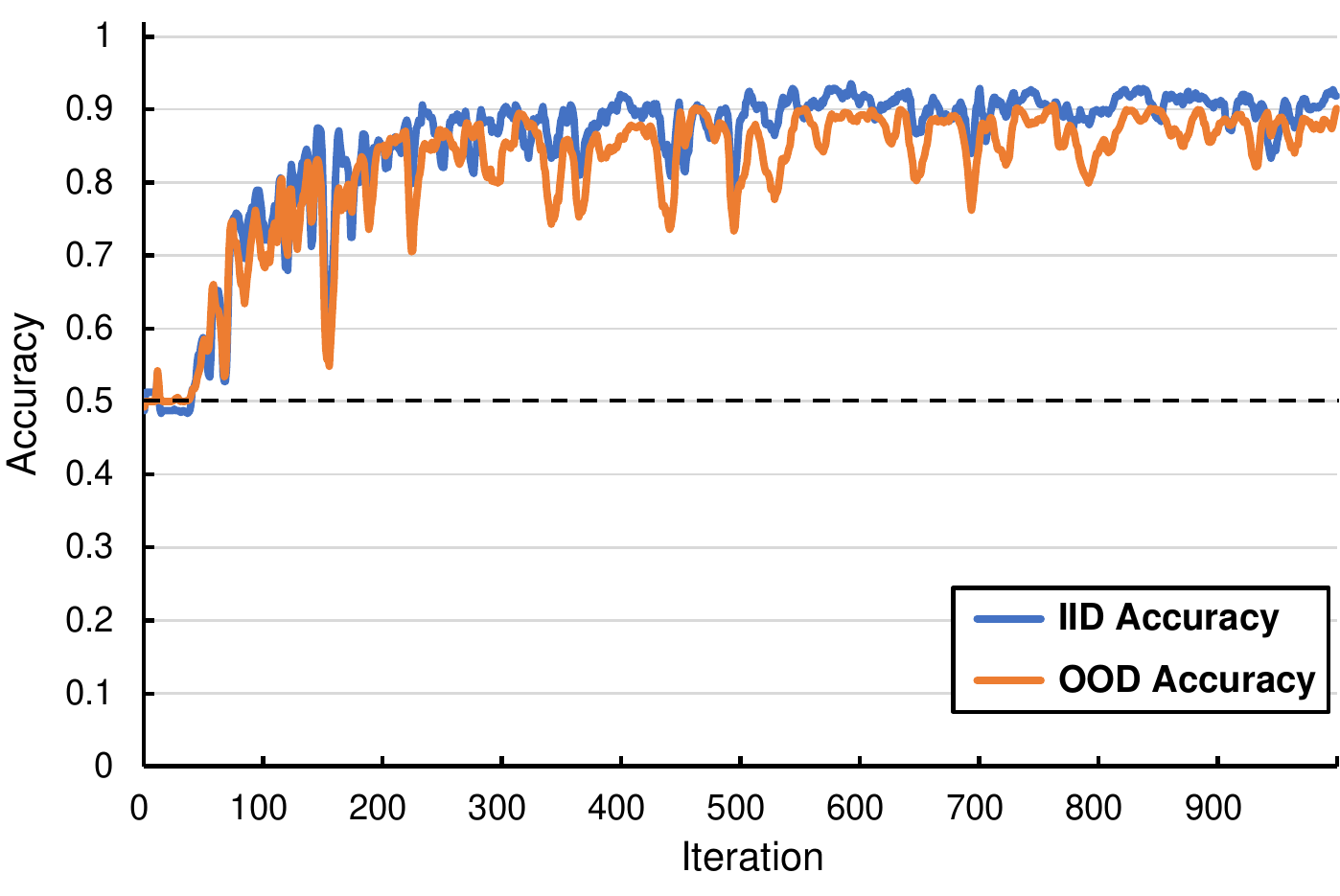}
        \caption{Marginal Spurious Correlation}
        \label{fig:iter-a}
    \end{subfigure}
    \begin{subfigure}{0.48\linewidth}
        \centering
        \includegraphics[width=0.8\linewidth]{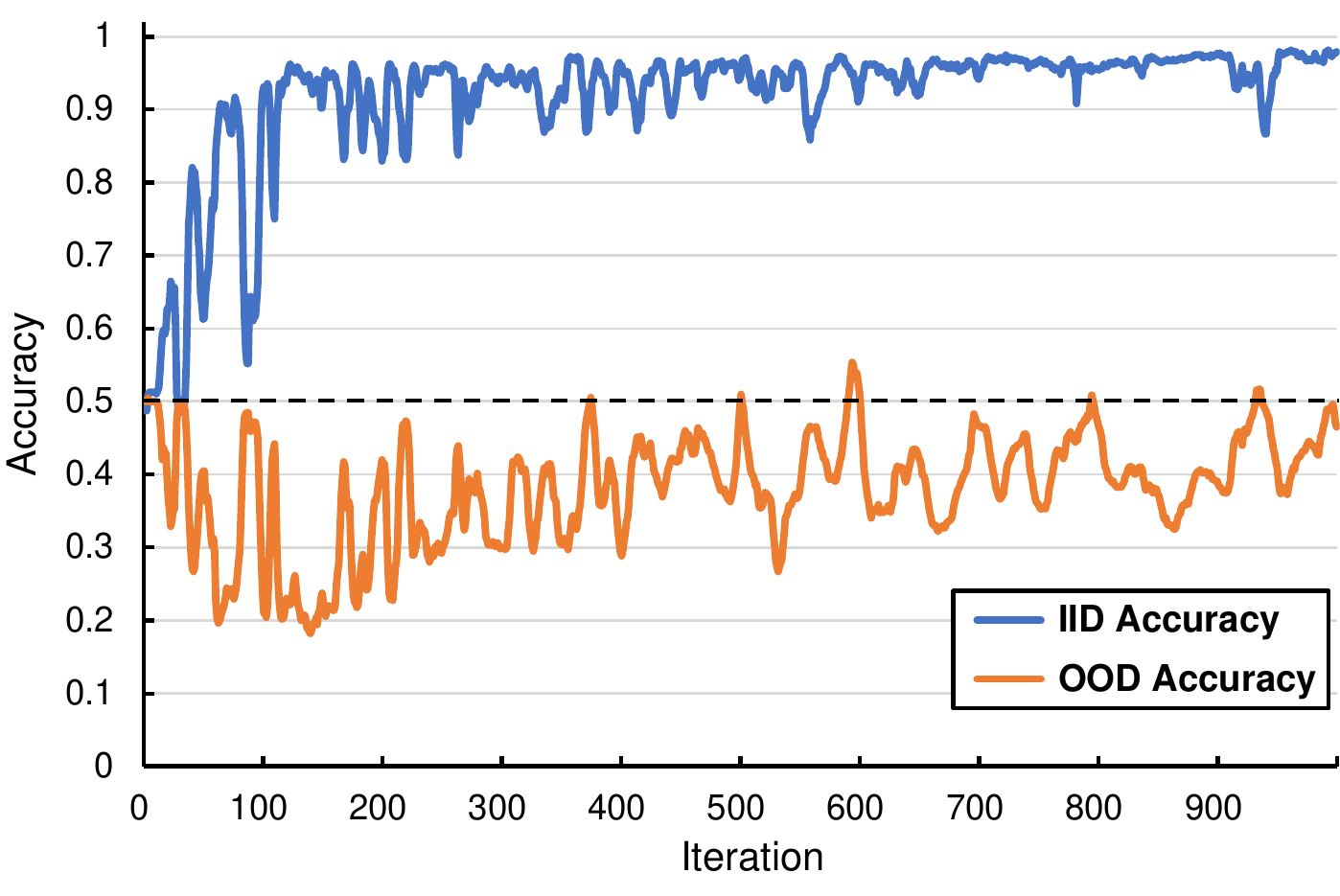}
        \caption{Conditional Spurious Correlation}
        \label{fig:iter-b}
    \end{subfigure}
    \caption{Comparison of the model accuracy between on the in-distribution validation set and on the out-of-distribution test set.}
    \label{fig:iter}
\end{figure*}

From the results in Table \ref{table:ood result}, it can be seen that the accuracy of models declines on the OOD test set with marginal distribution shift, which is not in line with human intuition. Intuitively, the models should not use an irrelevant feature as a discriminative feature when the irrelevant feature and label are not correlated in the training data. In other words the model performance should not be affected even if the distribution of the irrelevant feature shifts in the test set. We suspect the reason for the experimental phenomenon contradicting intuition is that the deep models have learned the marginal spurious correlation.

To verify this point, we construct two groups of data based on MNIST \cite{lecun1998gradient}. The first group randomly selects 1,000 images from MNIST and creates 4 copies of different colors: red, yellow, green and blue. So these four copies are consistent except for digits colors. The second group uses the full MNIST samples, and the color settings are the same as the first group. We train a ResNet18 model and a 2-layer CNN model on the above two groups of data, using one copy for training each time, and the other copies for testing. We report ID validation accuracy, OOD test accuracy and the average drop rate of the two models in Table \ref{table:marginal-sc}.

The results in Table \ref{table:marginal-sc} show that: 1) whether the model size is large or small, performance drops when training on one color and testing on other colors, indicating that the models have learned marginal spurious correlations; 2) as \cite{hendrycks2021many,TaoriDSCRS20robustness} find that diverse training data can improve models' robustness to image corruptions and dataset shift, we find the increase of training data reduce the models' learning of marginal spurious correlations. For example, the accuracy of the ResNet18 trained on \textit{R-Full} drops by an average of 20.4\% in other color settings, which is 1/3 of that trained on \textit{R-1K}; 3) while \cite{hendrycks2021many} suggests that larger models can reduce the gap between ID accuracy and OOD accuracy, we find that larger models are more likely to learn marginal spurious correlation than small models. For example, the average drop rate of the 2-layer CNN model on OOD test sets is much smaller than that of ResNet18.

To explain why the model exhibits such counter-intuitive behavior, part of the clues already exist in \cite{soudry2018implicit, NagarajanAN21understanding}: the gradient descent-based optimization algorithm used for model training can have a slow logarithmic rate of convergence, and it usually cannot find the optimal parameters. Even if an irrelevant feature does not correlate with the label in the data, the model trained by gradient descent can unintentionally use it as a discriminative feature. See illustration in Figure \ref{fig:ds-b}.

\subsection{Conditional Spurious Correlation}

\begin{figure*}[!tb]
  \centering
  \begin{subfigure}{0.47\linewidth}
    \centering
    \includegraphics[width=1\linewidth]{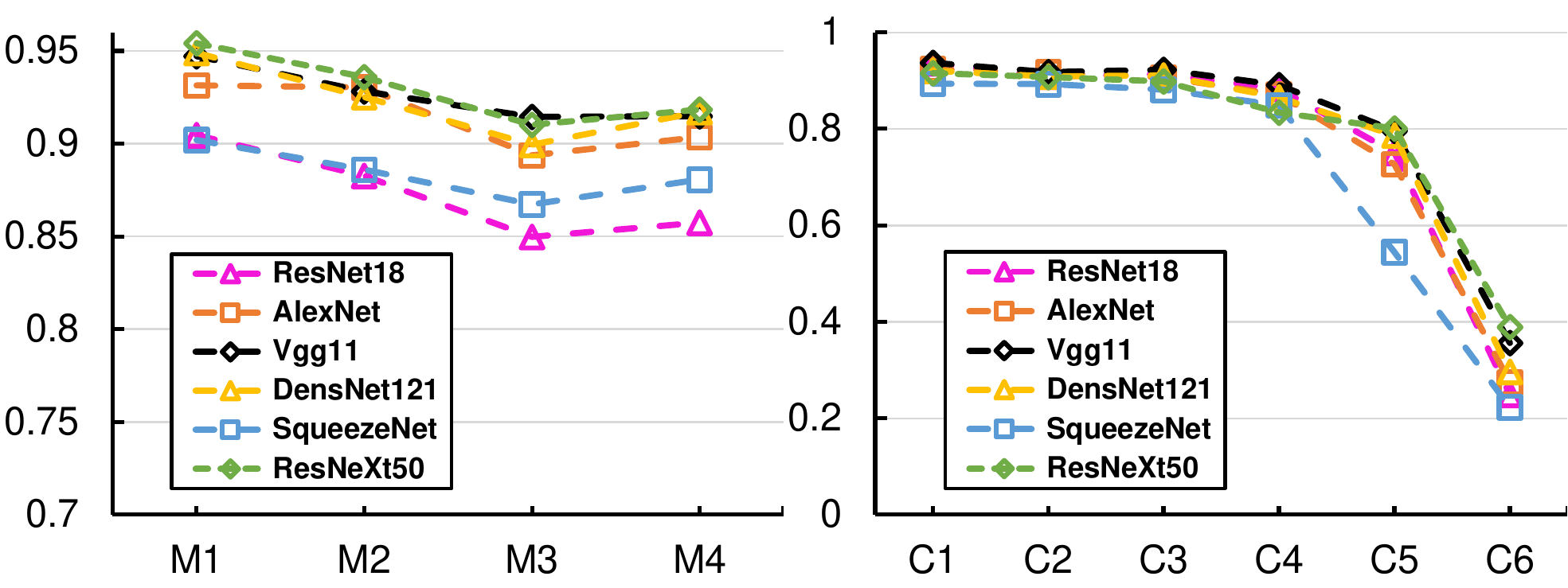}
    \caption{CelebA}
    \label{fig:level-celeba}
  \end{subfigure}
  \hfill
  \begin{subfigure}{0.47\linewidth}
    \includegraphics[width=1\linewidth]{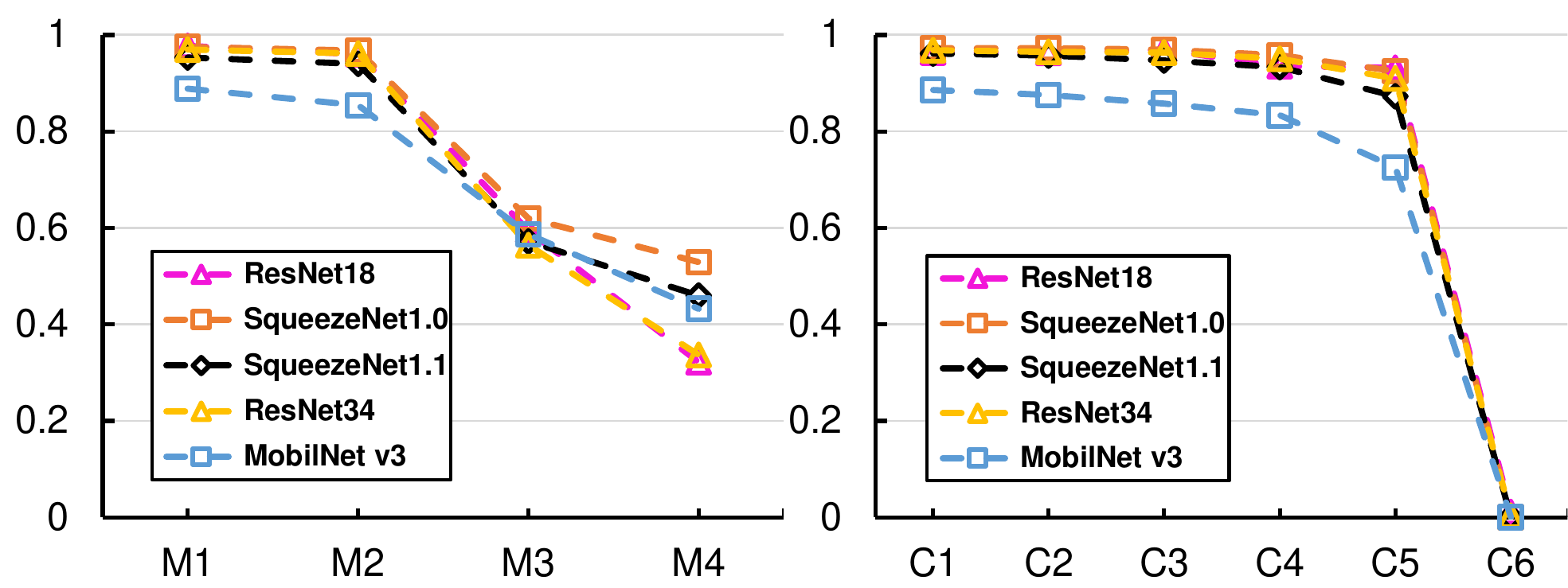}
    \caption{Colored MNIST}
    \label{fig:level-mnist}
  \end{subfigure}
  \caption{Performance of models under different levels of distribution shift. We construct 4 levels of marginal distribution shift, $M1$ \textasciitilde $M4$, by varying the proportion of hair colors (or digits colors) in the datasets; and six levels of conditional distribution shift, $C1$ \textasciitilde $C6$, by changing the correlation degrees between hair colors (or digits colors) and labels.}
  \label{fig:level}
\end{figure*}

The models are greatly affected by the conditional distribution shift. For example, ResNet18 can achieve the accuracy of 99.36\% on ID validation set of CelebA but only the accuracy of 27\% on the OOD test set. We consider that the model intentionally learns the conditional spurious correlation employing incorrect feature-label correlations in the training set, and relies heavily on it when testing.
To further explore how conditional spurious correlation affects model performance and how it differs from marginal spurious correlation, we report the accuracy for the first 1,000 iterations of ResNet18 with a batch size of 16 in two OOD settings (marginal and conditional distribution shifts). As shown in Figure \ref{fig:iter-b}, the model shows a polarization trend at the very beginning, and achieves ID accuracy of 95\% and OOD accuracy of 20\% in the early iterations. This demonstrates that in such an OOD setting, the model learns almost only conditional spurious correlations in early training. In contrast, Figure \ref{fig:iter-a} shows there is a small gap between ID accuracy and OOD accuracy in the later iterations of training which indicates the model has a slower learning rate for marginal spurious correlation. Further, in Figure \ref{fig:iter-b}, after about 200 iterations, the ID accuracy of the model stabilizes and there is a small rebound in OOD accuracy which always stays below 50\%. This indicates that as the training progresses, the model can slightly reduces the use of conditional spurious correlations.

%%%%%%%%%%%%%%%%%%%%%%%%%%%%%%%%%%%%%%%%%%%%%%%%%%%%%%%%%%%%%%%%%%%%%%%%%%
\section{Towards OOD Test}
\label{sec:towards}

% \begin{figure}[!tb]
%     \centering
%     \includegraphics[width=1\linewidth]{bug.pdf}
%     \caption{GradCAM of ResNet18 on CelebA. The first column is the original image, and columns from 2 to 7 show the activation of the model on the images when trained at different levels of spurious correlation.}
%     \label{fig:bug}
% \end{figure}

While a lot of works are devoted to improving the OOD generalization of models, there is still a great gap in how to evaluate the model performance in handling unseen OOD data. In this section, we propose OOD test paradigms for both scenarios when irrelevant features are known and unknown, using different OOD data settings, and we discuss how to use OOD test results to find bugs in models and guide debugging.

%%%%%%%%%%%%%%%%%%%%%%%%%%%%%%%%%%%%%%%%%%%%%%%%%%%%%%%%%%%%%%%%%%%%%%%%%%
\subsection{Evaluation with Different Levels and Types of OOD Test Set}
\label{sec:different}

Since the models can be affected by spurious correlations during training (see Section \ref{sec:reasons}), we cannot use ID performance to measure the real OOD performance of the model. Although some prior works have already recognized the importance of OOD test \cite{agrawal2018don,AkulaGAZR20words,BarbuMALWGTK19object,guo2019quantifying,TeneyAKSKH20value}, their test methods are incomplete, such as using only one single OOD test set for evaluation. As distribution shifts in practical scenarios are diverse, we suggest constructing OOD test sets with different levels and types of distribution shift to more comprehensively and accurately evaluate the OOD performance of the models.

To prove the effectiveness, we construct two types and multiple levels distribution shifts on CelebA and Colored MNIST to evaluate the OOD performance of several models. As shown in Figure \ref{fig:level}, as the degree of distribution shifts increases, the performance of the models is basically declining. Specifically, as the degree of marginal distribution shifts increases, the model performance decreases at a nearly linear trend, indicating that the model is progressively more influenced by the marginal spurious correlation. However, as the degree of conditional distribution shifts increases, the model is less affected at the beginning and has a sharp drop at the late stage indicates that the model is strongly influenced by conditional spurious correlations in extreme settings. With these results, we can infer the possible performance of these models in practical scenarios, helping us to anticipate the possible risks.

%%%%%%%%%%%%%%%%%%%%%%%%%%%%%%%%%%%%%%%%%%%%%%%%%%%%%%%%%%%%%%%%%%%%%%%%%%
\subsection{Evaluation with Clustered OOD Test Set}

\begin{table}[!t]
\centering
\caption{Comparison of accuracy of ResNet18 and SqueezeNet on clustered Colored MNIST.}
\resizebox{1\linewidth}{!}{
\begin{tabular}{lrrrr}
\toprule
   & \multicolumn{2}{c}{ID Acc} & \multicolumn{2}{c}{OOD Acc (Avg)} \\
   \cline{2-3} \cline{4-5}
   & \small{ResNet18} & \small{SqueezeNet} & \small{ResNet18}  & \small{SqueezeNet}       \\
   \hline \hline
D1 & 96.06  & 96.62     & 94.37            & \textbf{95.59}            \\
D2 & 94.56  & 96.30     & 87.22            & \textbf{93.91}            \\
D3 & 97.75  & 96.77     & \textbf{43.19}   & 43.17            \\
D4 & 96.59  & 96.42     & 37.79            & \textbf{42.40}             \\
D5 & 95.39  & 95.39     & 41.65            & \textbf{50.99}            \\
D6 & 94.67  & 94.67     & 56.17            & \textbf{59.14}            \\
D7 & 91.84  & 94.56     & 43.60            & \textbf{61.75}            \\
D8 & 79.07  & 86.05     & 34.86            & \textbf{43.14}            \\
\bottomrule
\end{tabular}
}
\label{table:cluster}
\end{table}

\begin{figure*}[!t]
    \centering
    \includegraphics[width=0.9\linewidth]{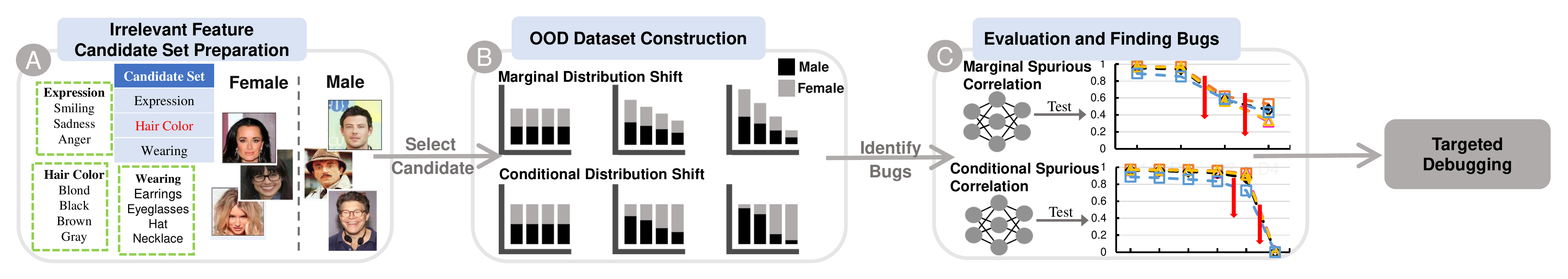}
    \caption{Employing OOD test to find bugs for deep network models.}
    \label{fig:framework}
\end{figure*}

Knowing the labels and specific irrelevant features of the data (e.g., NICO \cite{he2021towards} gives both labels and backgrounds of the images), we can directly use the features to construct different types and levels of OOD test sets. But in general, we have only the raw data without irrelevant features that can be used directly. In such a scenario, we suggest clustering the data to construct OOD test sets with implicit distribution shifts.

We use a ResNet50 pre-trained at ImageNet as the feature extractor, and the input vectors of the fully-connected layer as the embedding representation of samples. We then cluster the samples of each category of Colored MNIST into 8 clusters in the embedding space, and combine them in order of cluster size to obtain 8 sub-datasets $(D_i)_i^{N=8}$. We use one for training and the others for testing, and report ID accuracy and average OOD test accuracy in Table \ref{table:cluster}. The results show that different clustered OOD test sets bring different degrees of performance decline which demonstrates that clustered OOD data can evaluate the OOD performance of the models to some extent. And we can use these results to compare models to choose the better one. For example, a small model seems a better choice than a large model for the problem of colored handwritten digit recognition.

%%%%%%%%%%%%%%%%%%%%%%%%%%%%%%%%%%%%%%%%%%%%%%%%%%%%%%%%%%%%%%%%%%%%%%%%%%

\subsection{Discussions on Finding Bugs for Deep Network Models}

Similar to software engineering, finding ``bugs'' in machine learning, i.e., which irrelevant features significantly undermine the performance, plays a critical role in testing before the model is deployed to production  \cite{berend2020cats}. In current machine learning test, however, finding bugs is largely ignored. One reason is that, machine learning models, especially deep network models conduct end-to-end inference as a black-box fashion, making traditional testing paradigm such as unit testing in software engineering failed to be directly applied. In this section, we discuss the possibility of employing the proposed OOD test paradigm to identify the significantly harmful irrelevant features. The following elaborates the three steps of finding bugs for deep network models. This is illustrated in Figure \ref{fig:framework}.

\noindent \textbf{Irrelevant Feature Candidate Set Preparation.} The workflow starts with preparing a candidate set of all possible irrelevant features $(\widetilde{x}_i)_i^n \subseteq \widetilde{X}$ for the examined task. For example, for the task of image gender classification, we may construct the candidate set consisting of irrelevant features concerning \emph{experssion, hair color, wearing,} etc.

\noindent \textbf{OOD Dataset Construction.} Given one irrelevant feature to be tested, e.g., $\widetilde{x}_i=\mbox{hair color}$, we then construct OOD datasets by modifying the distribution of $\widetilde{x}_i$ with different types and levels~\footnote{This requires the access to irrelevant feature annotation in the test data, i.e., the hair color label for each sample. For general case when the irrelvant feature annotation is not available, we need to prepare a light irrelevant feature classifier before OOD dataset construction.}.

\noindent \textbf{Evaluation and Finding Bugs.} Evaluation can be conducted on the above constructed different settings of OOD datasets. By examining the performance decline in different types/levels of distribution shifts, we can identify bugs in two levels: (1) whether the tested irrelevant feature is bug (i.e., siginicantly undermines the performance), (2) which type of spurious correlation the model learned and mainly relied on.

The above workflow iterates till we test all the candidate irrelevant features to derive the identified bugs for current deep network model. This enables targeted debugging and developing corresponding solutions for more effective model improvement. For example, the identified specific irrelevant feature (e.g., \emph{hair color}) leads to data augmentation on balancing the distribution or model regularization on penalizing features regarding \emph{hair color}.

\section{Related Work}

As models can significantly fail when the test distribution is distinct from the training distribution \cite{recht2019imagenet,TaoriDSCRS20robustness,hendrycks2019benchmarking,hendrycks2021natural}, a range of advanced algorithms are proposed for OOD generalization, such as invariant representation learning \cite{wang2020unseen,blanchard2011generalizing,ghifary2015domain,ganin2015unsupervised}, feature disentanglement \cite{li2017deeper,xu2014exploiting,niu2015multi}, adversarial data augmentation \cite{shankar2018generalizing,volpi2018generalizing,zhou2020deep}. While these algorithms improve the OOD generalization of models under their own experimental settings, a recent work \cite{gulrajani2020search} finds that these algorithms may be overestimated, i.e., most advanced OOD generalization algorithms are only comparable with Empirical Risk Minimization (ERM) \cite{vapnik1998statistical} when under the same experimental settings. Therefore, in addition to developing algorithms that improve the OOD generalization, it is also important to develop methods that can better evaluate the capability of algorithms to handle unseen OOD data.

A number of benchmarks have been introduced to develop OOD generalization in both CV \cite{hendrycks2019benchmarking, he2021towards,li2017deeper,fang2013unbiased} and NLP \cite{hendrycks2020pretrained,miller2020effect}, but most of these benchmarks build only one type of distribution shift. DomainBed proposed by \cite{gulrajani2020search} includes several OOD generalization datasets, so it has a richer type of distribution shift, but not multi-level distribution shifts. This suggests that we need a new testing paradigm to test the OOD generalization capability of the models.

\section{Conclusion}

% In this work, we categorize distribution shift into three types: marginal, conditional and joint distribution shift, and find that ID test results cannot reflect models' real performance in practical implementation and also cannot represent relative ranks. Experiments results show that models learn two different spurious correlations in corresponding OOD data settings so as to make ID test suffers the above problems. Based on this, we suggest using different levels and types of OOD test sets to evaluate performance of models in pratical scenario. Following the OOD test results, we find bugs of models by identifying what features impair model performance most, and expect a guideline for model debugging will be provided in the future works.

% In this work,To accomplish this, we first introduce a unified perspective of distribution shift between the observed training data and unseen test data, and find the problem of ID test: the ID performance cannot represent the OOD performance. We then ascribe the reason of IID test failures as marginal spurious correlation and conditional spurious correlation.

We propose practical OOD test paradigms to evaluate the OOD performance of the model and a workflow for finding bugs that can bring guidelines for model debugging. To accomplish this, we first introduce a unified categorization perspective of distribution shift between the observed training data and unseen test data, and find the problem of ID test: the ID performance cannot represent the OOD performance. We then ascribe the reason for IID test failures to marginal and conditional spurious correlation. We expect the test paradigms and workflow can lead to more focusing on the testing of models' capabilities to handle OOD data.

%%%%%%%%% REFERENCES %%%%%%%%%%%%%%%%%%%%%%%%%%%%%%%%%%%%%%%%%%%%%%%%%%%%%

{\small
\bibliographystyle{ieee_Fullname}
\bibliography{egbib}
}

\end{document}